\documentclass[runningheads]{llncs}
\usepackage[T1]{fontenc}
\usepackage{graphicx}
\usepackage[utf8]{inputenc}
\usepackage{caption}
\usepackage{amssymb}
\usepackage{mathtools}
\usepackage{hyperref}
\usepackage{todonotes}
\usepackage{wrapfig}

\usepackage{booktabs}
\usepackage{multirow}
\usepackage{adjustbox}
\usepackage{colortbl}

\begin{document}
\title{LogRCA: Log-based Root Cause Analysis for Distributed Services}
%
%
\author{Thorsten Wittkopp \and
Philipp Wiesner \and
Odej Kao}
%
\authorrunning{T.Wittkopp et al.}
%
\institute{Technische Universität Berlin\\
\email{\{t.wittkopp, wiesner, odej.kao\}@tu-berlin.de}}
\maketitle              
\begin{abstract}

To assist IT service developers and operators in managing their increasingly complex service landscapes, there is a growing effort to leverage artificial intelligence in operations.
To speed up troubleshooting, log anomaly detection has received much attention in particular, dealing with the identification of log events that indicate the reasons for a system failure.
However, faults often propagate extensively within systems, which can result in a large number of anomalies being detected by existing approaches.
In this case, it can remain very challenging for users to quickly identify the actual root cause of a failure.

We propose LogRCA, a novel method for identifying a minimal set of log lines that together describe a root cause.
LogRCA uses a semi-supervised learning approach to deal with rare and unknown errors and is designed to handle noisy data.
We evaluated our approach on a large-scale production log data set of 44.3 million log lines, which contains 80 failures, whose root causes were labeled by experts.
LogRCA consistently outperforms baselines based on deep learning and statistical analysis in terms of precision and recall to detect candidate root causes. 
In addition, we investigated the impact of our deployed data balancing approach, demonstrating that it considerably improves performance on rare failures. 
\end{abstract}


\section{Introduction}



The complexity of modern IT services presents operation and development teams with challenges in terms of both implementation and maintenance~\cite{rosendo2018improve}. 
To address this issue, approaches in artificial intelligence for IT operations (AIOps) aim to support users in troubleshooting and problem mitigation~\cite{gulenko2020ai}.
A key use case for AIOps is to detect and resolve system failures and associated root causes. 
Due to the heavy use of logging in modern systems, many works have hereby focused on log anomaly detection~\cite{zawawy2010log,bogatinovski2020multi,wittkopp2021loglab,DBLP:conf/hicss/WittkoppSWAK23,hamooni2016logmine,korzeniowski2022landscape,lu2017log}.

However, faults often propagate extensively within systems before actual failures occur.
This can result in large amounts of anomalies being detected across various different services.
Presenting users with hundreds of potentially anomalous log lines is not a sensible approach when the main concern is to quickly understand the root cause of a failure, and then mitigate it in further updates.

In complex service architectures, the root cause of a failure is often not determined by a single event.
Instead, it must be described by a set of faults and system states that are often distributed across many different services.
These services generate large amounts of log data, which is why users can easily overlook important information.
The key challenge in \emph{root cause analysis} is to identify the minimal set of relevant information that is required to understand the root of a system failure.
Several recent works explore root cause analysis for log data~\cite{korzeniowski2022landscape}.
However, a significant drawback of these studies is that they were developed for specific systems, where the number of possible root causes was predetermined~\cite{lu2017log,lu2019ladra,sharp2016semi}. 
This limitation makes it difficult to apply these methods in complex and constantly evolving systems. 
In most real-world services, not all root causes are known in advance. 

To facilitate the quick understanding of failures in distributed IT services, we propose a log-based root cause analysis method which is trained in a semi-supervised fashion.
Our approach, called LogRCA, determines a set of log lines that together describe the root cause of a failure.
For this, LogRCA automatically ranks all log lines that were generated within a \emph{investigation time window} before the failure by relevance.
Users can then dynamically investigate different thresholds, resulting in a set of \emph{root cause candidates}.
These root cause candidates are expected to be temporarily ordered and causally related~\cite{9529498}.
The contributions of this paper are the following:
\begin{itemize}
\item We propose a method for identifying a set of log lines that are describing the root cause of a system failure. Our method utilizes semi-supervised learning and is based on a transformer model with custom objective function and can handle very noisy data.
\item We propose an approach to improve performance on rare and unknown failures by balancing training data before training the model.
\item We evaluated our approach on a large-scale production log dataset produced by 46666 different services, where the root cause log lines of 80 system failures have been labeled by experts.
\end{itemize}

Section~\ref{sec:towards} explains the problem and challenges.
Section~\ref{sec:PU_learning} describes how we formulate root cause analysis as a PU learning problem.
Section~\ref{sec:balancing} proposes a data balancing approach to improve performance on rare root causes.
Section~\ref{sec:approach} presents our root cause analysis approach in detail.
Section~\ref{sec:evaluation} evaluates LogRCA against different baselines and investigates the impact of data balancing.
Section~\ref{sec:related-work} surveys the related work.
Section~\ref{sec:conclusion} concludes the paper.

\section{From Anomaly Detection to Root Cause Analysis}
\label{sec:towards}
\begin{figure}[htbp]
\centering

\includegraphics[width=0.8\columnwidth]{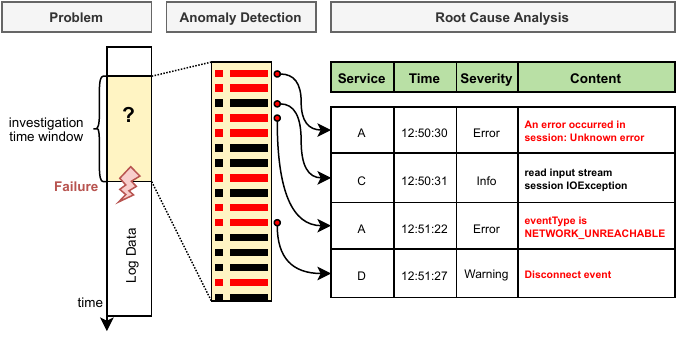}
\caption{LogRCA helps users to identify a minimal set of root cause log lines that reside within a investigation time window prior to the failure.}
\label{fig:problem_description}
\end{figure}

As failures can propagate through services in complex distributed systems, traditional anomaly detection approaches often alert for very large numbers of anomalous log lines, limiting their usefulness to users.
The main difference between log anomaly detection and log-based root cause analysis is that anomaly detection aims to identify anomalous behavior in IT services by selecting individual (and often contextually unrelated) log lines. Log-based root cause analysis, on the other hand, aims to select a minimal set of contextually and timely related log lines to support development or operations teams in better understanding the actual root cause of a failure.
This is a nontrivial task, since the aim of root cause analysis is to give insight into the actual course of action~\cite{zawawy2010log} of the failure.
For example, there might be important log lines that help a team understand the root cause, which would not have been detected by traditional anomaly detection because they describe normal behavior.
Figure \ref{fig:problem_description} exemplifies this problem and the desired solution. 

Due to the lack of information about the characteristics of a root cause, the process of root cause analysis poses three main challenges:

\begin{enumerate}
    \item We do not know which log lines within an investigation time window represent the root cause of a failure. This means that we must treat all log lines prior to a failure as possible root cause candidates, which leads to a large number of incorrectly labeled data during training.
    \item The number of log lines representing a root cause can differ from case to case.
    In contrast to the binary classification required in traditional anomaly detection, our goal is to identify an unknown number of log lines. 
    \item Training data is often very unbalanced, which means that some root causes have been common in the past, while others have rarely or never occurred.
    This can harm model performance due to training biases.
\end{enumerate}

\section{Root Cause Analysis as a PU Learning Problem}
\label{sec:PU_learning}
\begin{figure}[htbp]
\centering
\includegraphics[width=.9\columnwidth]{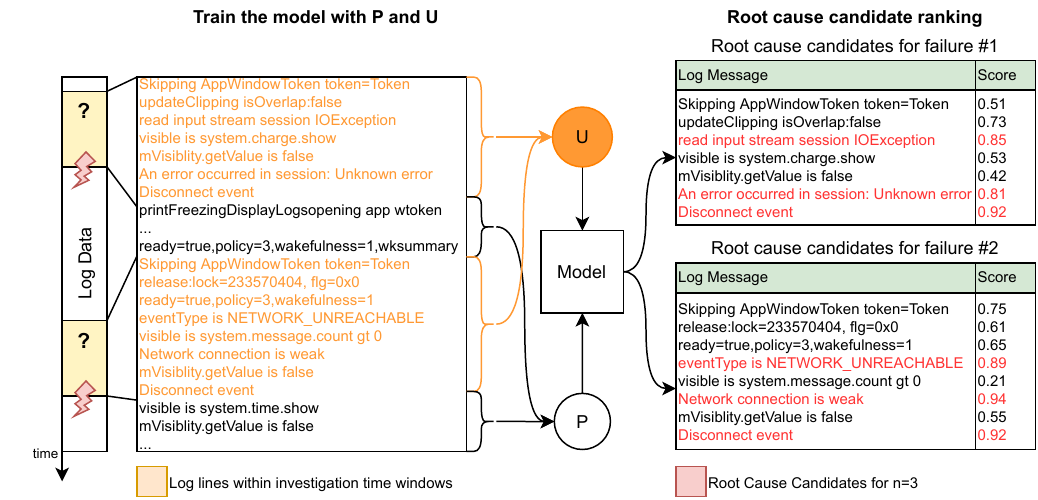}
\caption{Illustrating the training process with incorrectly labeled data and the result for $n=3$. Log lines in orange have been assigned to the unknown class $\mathcal{U}$, while black log lines are assigned to the normal class $\mathcal{P}$.}
\label{fig:problem_example}
\end{figure}

As a first step of our root cause analysis approach, we require the actual detection of a failure.
This can be determined through specific logs (for example, a dedicated \emph{disconnect} event like in our evaluation data) or through other monitoring systems.
Users then have to determine a reasonable \emph{investigation time window}, which describes the amount of time prior to the failure in which we expect the root cause log lines to reside.
The example in Figure~\ref{fig:problem_example} contains two windows (in yellow with a question mark) before their respective failures (red flash).

We formulate our training as a \emph{PU learning} problem~\cite{bekker2020learning}.
PU learning is an umbrella term for several binary classification methods that learn the distribution of positive samples (in our case, log lines outside the investigation time windows) to classify unknown samples~\cite{liu2002partially,zhu2009introduction,bekker2020learning,liu2003building}.
In other words, we train a machine learning model based on two classes: A \emph{positive} class $\mathcal{P}$ with normal data and an \emph{unknown} class $\mathcal{U}$, which contains both normal log lines and root cause log lines (marked orange in Figure~\ref{fig:problem_example}).
This leads to a training setup with a large number of inaccurate labels, as most samples in $\mathcal{U}$ are considered normal log lines.
As we do require information on when a failure occurred, as well as the investigation time window, we can also speak of a weakly supervised learning setting~\cite{zhou2018brief}.

After determining which log lines belong to $\mathcal{P}$ or $\mathcal{U}$, we train our model (explained in Section~\ref{sec:approach}).
Our model uses a custom objective function to assign each log line in $\mathcal{U}$ a \emph{root cause score} that determines the predicted relevance of the log line.
As we do not previously know how many log lines are required to fully understand a particular (and often entirely unknown) root cause, LogRCA does not decide the threshold of how many log lines should be presented itself.
Therefore, users are responsible to experiment with different thresholds, namely the number of log lines to be displayed, which correspond to the $n$ log lines with the highest scores in each investigation time window.





\section{Balancing Data to Boost Performance on Rare Cases}
\label{sec:balancing}

To improve our performance in rare or even unknown cases, we employ a training data balancing strategy based on automatic clustering.

\subsection{Imbalanced Training Data}

In practical settings, a particular system failure can be caused by a variety of different root cases.
For example, unexpected disconnect events of clients in a mobile computing setting can be caused by a variety of reasons, ranging from weak wireless network connection, over actual software crashes, to power supply issues.
However, some of these causes may be much more common than others, resulting in highly imbalanced training data.
If there are only very few samples of a specific root cause available within $\mathcal{U}$, we cannot expect the machine learning model deployed to sufficiently learn the distribution of the corresponding root cause log lines.
The bias towards majority classes will result in a model that may struggle to distinguish actual root cause log lines from log lines in the normal class $\mathcal{P}$ and hence assign a relatively low root cause score.

While handling class imbalance in deep learning is a well-researched problem~\cite{johnson2019class_imbalance_survey}, in our scenario we are operating in a semi-supervised setting.
Therefore, we neither know the class data distribution nor how many classes there are.
To deal with this, we \emph{estimate} the number of root causes by automatic clustering and then balance the data, with the goal of improving performance in underrepresented root causes.



\subsection{Balancing Through Automatic Clustering}

To balance the training data, we use automatic clustering to obtain an estimate of the number of root causes and their occurrences in the training data set.
Each cluster estimates a different type of root cause.
After the estimation, we balance the training data so that rare root causes are weighted stronger in the training process, although still not as strong as common root causes.

\begin{figure}[htbp]
\centering
\includegraphics[width=0.8\columnwidth]{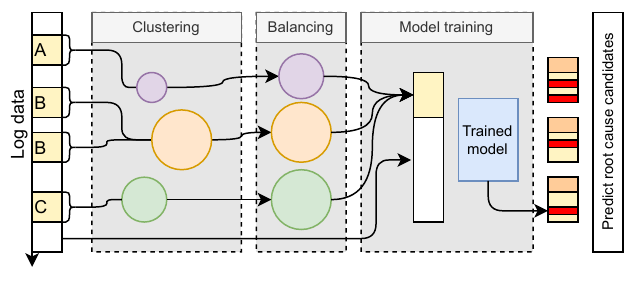}
\vspace{-5mm}
\caption{Balancing the training data.}
\label{fig:balancing}
\end{figure}

Figure~\ref{fig:balancing} illustrates the balancing procedure. 
As a first step, investigation time windows, depicted in yellow on the left, are encoded in a vector $\vec{w}$, by utilizing the meta-information of each log line in those windows. 
Specifically, the services $x_i$ are utilized, so each vector represents the services involved in each root cause.
Therefore, we create a vector $\vec{w}$, with dimensionality equal to the different values of $x_i$, which serves as input for the clustering:
$$dim(\vec{w}) = |\{\forall x_i \in \mathcal{L}: unique(x_i)\}|$$

Each cluster within the automatic clustering output estimates a specific root cause.
This step is not accurate but still gives a good estimation to balance the training data.
In our example, the root cause 'B' occurs two times and the root cause 'A' and 'C' occurs one time each.
The size of the circle illustrates the number of log lines within the corresponding investigation time windows of a specific cluster.
The log lines from the investigation windows of 'B' are therefore combined in one cluster.

During the balancing step, the number of log lines in each cluster is calculated: $\mathcal{K}$: $U = \{|k_i|:\forall k_i \in \mathcal{K}\}$, where $min(u)$ is the number of log lines in the smallest cluster and $max(U)$ is the number of log lines in the largest cluster.
The numbers of log lines for each cluster in $\mathcal{K}$ are then normalized between $\frac{max(U)}{2}$ and $max(U)$. 
This means that the smallest cluster will have half as many log lines as the largest cluster. 
The target size $t(\cdot)$ of each cluster $k_i$ is then calculated by the following equation:

\begin{equation}
t_{|k_i|} = \frac{|k_i| - min(U)}{max(U) - min(U)} \cdot (max(U) - \frac{max(U)}{2}) + \frac{max(U)}{2})
\end{equation}

In this formula, $|k_i|$ is the number of log lines in each cluster, we want to normalize between the desired range of $\frac{max(U)}{2}$ and $max(U)$.
The smallest cluster is represented by $min(U)$ and the largest cluster by $max(U)$.

Overall, balancing the training data is a critical step, as it allows the model to develop comprehensive training of each possible root cause but still train properly for the most occurring root causes, since they should also be the ones that occur very often in general. 
Subsequently, the class $\mathcal{U}$ is now balanced, whereas the class $\mathcal{P}$ is not touched during this step.

\section{LogRCA}
\label{sec:approach}

\begin{figure}[htbp]
\centering
\includegraphics[width=0.8\columnwidth]{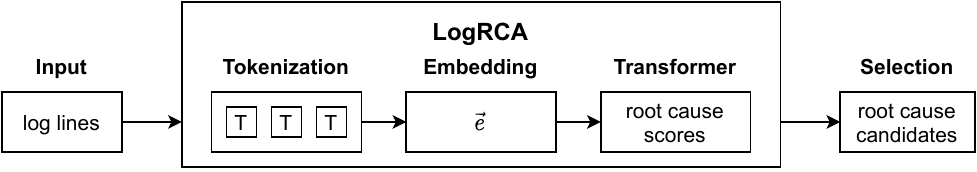}
\vspace{-1mm}
\caption{Steps for selecting root cause candidates.}
\label{fig:process}
\end{figure}

In this section, we present an approach for detecting a set of root cause log lines by examining an investigation time window prior to a failure through PU learning.
Our approach allows us to even identify the causes of rare or entirely new failures.
We propose a transformer-based neural network architecture with a custom loss function. 
Figure~\ref{fig:process} illustrates the different steps: After data preprocessing, the transformer model -- trained using PU learning -- determines a root cause score for each log line.
Users can then dynamically determine the number~$n$ of how many log lines should be presented for analysis.

\subsubsection{Input}
We define $\mathcal{L} = (l_i : i = 1,2, \ldots, n)$ as the set of all log lines.
Each log event $l_i \in \mathcal{L}$ can be decomposed into meta-information $m_i$ and content $c_i$, where the content serves as the input to the model. Therefore, the meta-information contains the service $x_i$ by which the log line was produced.

\subsubsection{Tokenization}
As a first step in preprocessing the content of log messages, we employ tokenization, a commonly used transformation from natural language processing.
Tokenization breaks down the content into its smallest indecomposable units, called tokens. Thus, each log content $c_i$ can be represented as a sequence $c_i = (w_j: w_j \in V, j = 1,2,\ldots, s_i)$ of tokens, where $V$ is the set of all known tokens, and $s_i$ denotes the total number of tokens in $c_i$.

Initially, the content $c_i$ of each log line $l_i$ is converted into a sequence of tokens $c_i$ using the separators: \texttt{,} and \texttt{:} and \texttt{whitespace}. 
The resulting sequence of tokens is then further processed by replacing certain tokens with placeholders that adequately represent the original token while preserving relevant information. 
A placeholder token \texttt{[IP]} is introduced for IP address values, \texttt{[NUM]} is used for any number greater or equal to 10, \texttt{[HEX]} is utilized for hexadecimal numbers and \texttt{[ADDR]} for internal addresses of the application.
Finally, the transformed sequence of tokens is prefixed with a special token \texttt{[CLS]} to encode the information of each log line later.
An exemplary log message
\begin{center}
    \footnotesize
    \texttt{Network ip: 192.168.0.1 weak connection}
\end{center}
is thus transformed into a sequence of tokens
\begin{center}
    \footnotesize
    \texttt{[[CLS], Network, ip, [IP], weak, connection]}.
\end{center}

Lastly, we truncate the token sequences obtained from the log events to a fixed length $s$ and pad smaller sequences with the special token \texttt{[PAD]} to match the required input size.

\subsubsection{Embedding}
An embedding $\vec{e_i}$ is a vector representation of a token to be used as input for a machine learning model.
A transformation function $g$ transforms a sequence of tokens $c_i$ with length $s_i$ into a sequence of embeddings $\vec{e_i}$, with $g: V^{|s_i|} \rightarrow \mathbb{R}^{d,|s_i|}$.
The embeddings are continuously adapted during the model training process to represent the semantics of the original token or sequence of tokens.
The $j$-th embedding in a sequence of embeddings $\vec{e_i}$ is denoted as $\vec{e_i}(j)$. 
Therefore, we compute an embedding vector $\vec{e}_i(j)$ for each token $w_j$ in the token sequence $c_i$. 
The resulting truncated sequences of embeddings $\vec{e}_i'$ are used as input to the neural network.


\subsubsection{Transformer}
To determine the root cause score for all log lines within an investigation time window, we utilize an encoder architecture with self-attention. 
As the network is expected to output a root cause score for each log line, which users later use to filter root cause candidates, we deploy a custom objective function.
This objective function must ensure that log lines in $\mathcal{U}$ receive scores if they are significantly different from log lines observed in $\mathcal{P}$.
On the other hand, the objective function should assign low scores to log lines that occur in both $\mathcal{P}$ and $\mathcal{U}$. 
Furthermore, the objective function should be able to handle a large number of mislabeled log lines that, by design, occur in the unknown class $\mathcal{U}$. 

To fulfill these requirements, we calculate scores based on the Euclidean distance, representing the length of the output vector $\lVert z_i\rVert$ for each input sequence $\vec{e_i}'$. 
The objective function comprises two parts: the first part minimizes errors for samples in class $\mathcal{P}$ to yield small scores close to zero, while the second part amplifies errors for samples in class $\mathcal{U}$ to drive higher scores. 
This structure is depicted in Equation \ref{eq:objective_function_gen}, where $\tilde{y}_i$ denotes the inaccurate label, $z_i$ represents the model output vector for each embedded input log message $\vec{e_i}'$, and $m$ indicates the number of samples per batch.

\begin{equation}
    \label{eq:objective_function_gen}
    \frac{1}{m}\sum\limits_{i=1}^{m}((1-\tilde{y}_i)*a(z_i) + (\tilde{y}_i)*b(z_i)
\end{equation}

For $a$, we minimize the error for positive samples and, in contrast, increase the error for all scores, when the log message is of class $\mathcal{U}$, with $a(z_i) = \lVert z_i \rVert^2$ and $b(z_i) = q^2 / \lVert z_i \rVert$, where $q$ is a numerator between 0 and 1 that represents the relation of the number of samples in $\mathcal{P}$ and $\mathcal{U}$.
Thus, the final objective function is composed as
\begin{equation}
    \frac{1}{m}\sum\limits_{i=1}^{n}\Big((1-y)*\lVert z_i \rVert^2 + (y)*\frac{(\frac{|\mathcal{P}|}{|\mathcal{P}|+|\mathcal{U}|})^2}{\lVert z_i \rVert}\Big)
\end{equation}

\noindent
This enables the transformer model to train log messages with inaccurate labels by modifying the calculated error based on the relation of $\mathcal{P}$ and $\mathcal{U}$.

\subsubsection{Selection}
As described in Section~\ref{sec:towards}, we cannot automatically determine how many log lines are required to fully understand a root cause.
Because of this, users must decide how many log lines should be returned for analysis.
Based on the previously calculated root cause scores, we then return the $n$ log lines with the highest scores in chronological order.
In practice, graphical user interfaces based on LogRCA can also make use of the root cause score to visualize the calculated relevance of a particular log line.
We evaluated the trade-off between precision and recall of different window sizes in Section~\ref{sec:evaluation}.

\section{Evaluation}
\label{sec:evaluation}

We evaluated LogRCA against different baseline approaches on a large-scale production dataset.

\subsection{Experimental Setup}

\subsubsection{Scenario and dataset.}
Our evaluation log data set was generated by 46666 different services running on Android devices. 
As the data originate from an industry production setting, it cannot be made available in this paper, but we describe its characteristics in this section.

\begin{wrapfigure}{r}{0.5\textwidth}
\vspace{-9mm}
\includegraphics[width=0.9\linewidth]{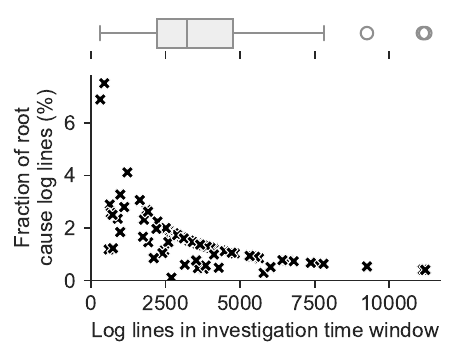} 
\vspace{-2mm}
\caption{Investigation time window sizes and their fraction of root cause log lines.}
\vspace{-2mm}
\label{fig:dataset}
\end{wrapfigure}

The data set comprises 44.3 million log lines (7.7 million unique), each consisting of 8.8 tokens on average.  
After replacing all tokens with the placeholders listed in Section~\ref{sec:approach}, we are left with 0.7 million unique log lines.

In total, the data covers 398 failures of Android devices.
For each failure, we define an investigation time window of 3 seconds prior to the failure, during which we expect to find the relevant log lines representing the root cause.
We consider 80 investigation time windows, where industry experts labeled the minimal set of root cause log lines for each failure.
The windows contain between 305 and 11209 log lines, each of which contains between 3 and 50 root cause log lines.
Figure~\ref{fig:dataset} shows the size distribution of the investigated time windows.

We train LogRCA and four baselines on the described log data. 
All log lines contained within the failure time windows ($\sim$300,000) are in the unknown class $\mathcal{U}$ and all remaining log lines ($\sim$44 million) in the positive class $\mathcal{P}$. 
For clustering, we used BIRCH~\cite{zhang1996birch} with branching factor $B=50$ and threshold $T=0.5$ to automatically determine the optimal number of clusters.





\subsubsection{Baselines.}

As described in Section~\ref{sec:related-work}, all the approaches reviewed for root cause analysis use a previously defined set of root causes.
LogRCA, on the other hand, aims to also help identify previously unseen root causes by assigning a root cause score to log lines and presenting the most relevant log lines to the user.
To evaluate our approach, we therefore train three statistical baselines and one neural network-based approach to assign a root cause score to log lines.

For all three statistical baselines, we employ TF-IDF (term frequency inverse document frequency) to encode tokens.
We utilize the predicted class probability as a score that determines the final ranking of root cause candidates.
First, a single \emph{Decision Tree} with a maximum depth of 30, where the score describes the fraction of samples of the same class in a leaf.
Second, a \emph{Random Forest} with 100 trees with a maximum depth of 20, where the score describes the mean predicted class probabilities of all trees in the forest.
Third, an \emph{SVM}, where the score is determined by a 5-fold cross-validation.
To make the SVM scores comparable, we calibrated its output probabilities using Platt scaling~\cite{Platt99}.

For neural network-based approaches, we trained a \emph{feedforward neural network (FNN)} and LogRCA for 5 epochs, where both networks comprise an embedding layer with 128 units to encode tokens.
The FNN baseline has two hidden layers with 256 hidden units each and its score is based on the calibrated output probability of its output layer~\cite{nn_calibration_2017}.
LogRCA's attention layer has two attention heads and connects to a fully connected layer with 256 units.
In these configurations, LogRCA and FNN take roughly the same time to train.

\subsection{Performance Analysis}

\begin{figure}[h]
\centering
\includegraphics[width=.9\columnwidth]{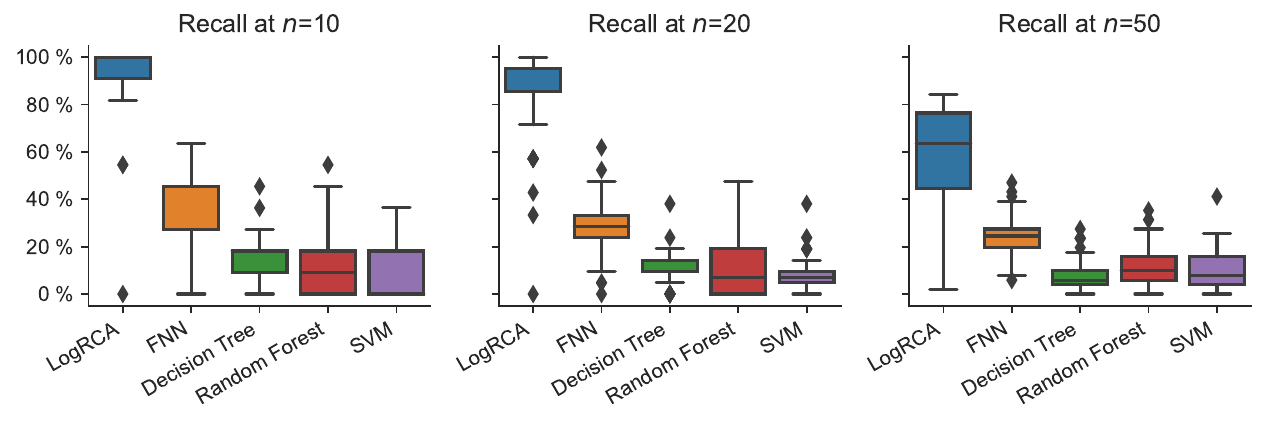}
\vspace{-3mm}
\caption{Fraction of root case log lines at 10/20/50 returned candidates.}
\label{fig:recall_at}
\end{figure}

We consider the recall of returned candidates to be the most relevant metric for users to quickly understand a root cause. \emph{We therefore want to investigate how many of the n highest ranked root cause candidates are actually part of the root cause.}
Note, that LogRCA does not automatically decide on a threshold~$n$, as we do not know how many log lines constitute the root cause in total.
Due to this, we analyze the recall at $n$=10, $n$=20, and $n$=50.
The results are presented in Figure~\ref{fig:recall_at}.

We observe that LogRCA clearly outperforms all baselines across the three windows.
Within the 10 top-ranked root cause candidates, LogRCA has an average recall of 93.5\,\%.
In 75 of 80 cases, all log lines are part of the root cause, and in the remaining 5 cases the root cause is fully covered as it consists of $\le$10 root cause log lines. To compare, the baseline FNN has an average recall of 36.1\,\% and the statistical approach only 15.8\,\% (Decision Tree), 10.7\,\% (Random Forest), down to 6.7\,\% (SVN).
Similarly, for 20 and 50 returned candidates, LogRCA maintains a high average recall of 86.6\,\% and 57.7\,\%, respectively, meaning that the majority of log lines presented to the user are actually part of the root cause.

This advantage also translates to other related metrics, such as the question of whether we actually identified \emph{all} root cause log lines.
Since we have up to 50 root cause log lines in our ground truth, we report this metric for $n$=50: 
In this case, LogRCA covered all root cause log lines in 65 out of 80 cases, compared to 57 for FFN, 29 for Decision Tree, 6 for Random Forest, and 11 for SVM.

\subsection{Impact of Balancing Training Data}

\begin{figure}[h]
\centering
\includegraphics[width=0.85\columnwidth]{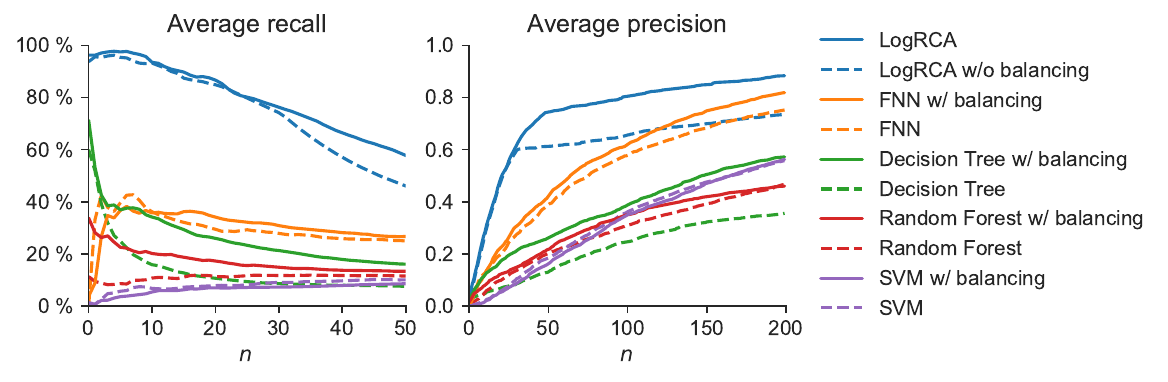}
\vspace{-3mm}
\caption{Average precision and recall of all approaches with and without balancing.}
\label{fig:precision_recall}
\end{figure}

To evaluate the impact of the proposed training data balancing, we apply this strategy to all baselines.
In return, we also trained a version of LogRCA without balancing.
Figure~\ref{fig:precision_recall} presents the resulting precision and recall of all approaches, with and without balancing.
We report recall until $n$=50 (since none of the investigation time windows contains more than 50 root cause log lines) and precision until $n$=200.

The intended effect of data balancing in LogRCA is to increase its performance on rare failures.
This is clearly visible: LogRCA's average precision and recall until $n \le 30$ independently of whether balancing is applied or not.
This suggests that many root causes were comparably ``easy'' for the transformer model to detect.
For $n > 30$, the effect of balancing considerably improves average performance, as there is less overfitting on frequently occurring failures.

Interestingly, balancing significantly improves the recall of the Decision Tree and Random Forest baselines, while there was no observable benefit to the SVM model.
For the FNN, balancing did not significantly improve or harm recall, but consistently improved precision with increasing root cause candidates.


\section{Related Work}
\label{sec:related-work}

\subsubsection{Log anomaly detection.}
In recent years, many log anomaly detection methods have been published to identify anomalous behavior in log datasets.
They focus on marking individual log lines as anomalous, but do not try to identify the minimal set of anomalies for each failure as a root cause.
LogCluster~\cite{LinZLZC16} is a clustering-based method that relies on log vectorization, clustering via Agglomerative Hierarchical Clustering, and extraction of cluster representatives.
DeepLog~\cite{du2017deeplog} uses templates~\cite{he2017drain} and an LSTM. 
By applying templating, it models logs as a sequence of events and performs anomaly detection on each.
LogAnomaly~\cite{MengLZZPLCZTSZ19} introduces the template2vec representation method and fuses it with LSTM networks into an end-to-end framework that detects sequential and quantitative anomalies. 
In~\cite{FarzadG20}, the authors propose the combination of Isolation Forests and multiple Autoencoder Networks. 
ADA~\cite{YuanALYL020} employs LSTM networks and dynamic thresholding to mark individual logs as anomalous.
LogClass~\cite{meng2021logclass} is a system designed to automatically and effectively recognize and categorize abnormal logs using partial labels. Integrates a technique for word representation, a positive and unlabeled learning (PU learning) model, and a machine learning classifier and uses Inverse Location Frequency (ILF) to accurately assign weights to the words in logs during feature generation.

\subsubsection{Root cause analysis.}

In the context of AIOps, root cause analysis aims at the identification of causal relationships between events to allow the identification of root causes of failures in IT system components.
For example, LogRule~\cite{notaro2023logrule} leverages structured logs and association rule mining to automate root cause analysis. 
The algorithm analyzes structured logs and generates a comprehensive list of explanations for a specific event.
The authors of~\cite{lu2017log} propose an offline approach that leverages Spark log files to accurately detect abnormalities and analyze their root causes. 
Their study considers four system resources: CPU, memory, network, and disk that could be the cause of a failure.
LogMine introduces a fast pattern recognition model for log analytics~\cite{hamooni2016logmine} method to extract patterns from a given set of log messages. 
LADRA~\cite{lu2019ladra} is a tool to detect abnormal data analytic tasks and perform root cause analysis using Spark logs. 
It employs a log parser to convert raw log files into structured data and extract relevant features. 
A detection method is then introduced to identify the occurrence time and location of abnormal tasks. 
Furthermore, predefined factors based on these features are extracted to facilitate root-cause analysis. 
General Regression Neural Network (GRNN) is employed to determine the likelihood of reported root causes, weighted by the factors.

All mentioned approaches aim to identify root causes from a previously known list of causes. In contrast, the goal of LogRCA is also to assist users in previously unseen root causes.

\subsubsection{PU learning.}

PU learning is frequently applied when it comes to anomaly detection.
LogClass~\cite{meng2018device} is a method for identifying and classifying anomalous logs in network and service management and combines word representation, PU learning, and a machine learning classifier. 
LogLAB~\cite{wittkopp2021loglab} proposes a modeling approach for the automated labeling of log anomalies without manual expert intervention, where the authors use information from monitoring systems to retrospectively generate precisely labeled data sets.
LogLAB also leverages the attention mechanism and a custom objective function tailored for weak supervision deep learning techniques.
The authors of PLELog~\cite{yang2021semi} propose a semi-supervised approach that minimizes manual labeling by using probabilistic label estimation based on historical anomalies. 
It incorporates an attention-based GRU neural network to efficiently detect anomalies while remaining stable with varying log data.
Lastly, PULL~\cite{DBLP:conf/hicss/WittkoppSWAK23} proposes an iterative log analysis method for reactive anomaly detection. 
Instead of relying on labeled data, it utilizes rough failure time estimations from monitoring systems to detect anomalies in log data.

\section{Conclusion}
\label{sec:conclusion}

%

We presented LogRCA, a new method for log-based root cause analysis in complex IT systems.
Our evaluation is based on a large-scale production log dataset labeled by experts.
We demonstrate how LogRCA has significantly higher recall than baseline methods and benefits from the proposed data balancing strategy for detecting the root cause of rare failures.

In the future, we want to investigate how many log lines are actually needed to provide sufficient context for users to understand a specific root cause.
Furthermore, we want to explore how the investigation time window size can be derived from the context of a failure automatically, instead of placing this decision on the user.

\bibliographystyle{splncs04}
\bibliography{bib}

\end{document}